\documentclass[10pt,twocolumn,letterpaper]{article}

\usepackage{wacv}
\usepackage{times}
\usepackage{epsfig}
\usepackage{graphicx}
\usepackage{subfigure}
\usepackage{amsmath}
\usepackage{amssymb}

\wacvfinalcopy 

\def\wacvPaperID{158} 

\ifwacvfinal\fi
\begin{document}

\title{Model-free Tracking with Deep Appearance and Motion Features Integration}

\author{Xiaolong Jiang$^{1,2,3}$ \hspace{1cm} Peizhao Li$^{1,2}$ \hspace{1cm} Xiantong Zhen$^{1,2,3}$ \hspace{1cm} Xianbin Cao$^{1,2,3}$ \\
$^{1}$School of Electronic and Information Engineering, Beihang University, China\\
$^{2}$Key Laboratory of Advanced technology of Near Space Information System, Beihang University,\\Ministry of Industry and Information Technology of China, China\\
$^{3}$Beijing Advanced Innovation Center for Big Data-Based Precision Medicine, China\\
\tt\small jasperj1tmac@163.com,{\tt\small \{lipeizhao,zhenxt,xbcao\}@buaa.edu.cn}
}
\maketitle
\ifwacvfinal\thispagestyle{empty}\fi

\begin{abstract}
Being able to track an anonymous object, a model-free tracker is comprehensively applicable regardless of the target type. However, designing such a generalized framework is challenged by the lack of object-oriented prior information. As one solution, a real-time model-free object tracking approach is designed in this work relying on Convolutional Neural Networks (CNNs). To overcome the object-centric information scarcity, both appearance and motion features are deeply integrated by the proposed AMNet, which is an end-to-end offline trained two-stream network. Between the two parallel streams, the ANet investigates appearance features with a multi-scale Siamese atrous CNN, enabling the tracking-by-matching strategy. The MNet achieves deep motion detection to localize anonymous moving objects by processing generic motion features. The final tracking result at each frame is generated by fusing the output response maps from both sub-networks. The proposed AMNet reports leading performance on both OTB and VOT benchmark datasets with favorable real-time processing speed.
\end{abstract}

\vspace{-2mm}
\section{Introduction}
\label{sec:Intro}
Visual object tracking has drawn intense research interests for decades with its vast implementations covering human-computer interaction, automatic driving, and visual surveillance, etc. Particularly, model-free tracking stands out considering its omnifaceted applicability. Being able to track any moving target in a sequence, a model-free tracker is readily plug-and-play without further pre-requests on the target. Nevertheless, the design of such a tracker is challenged by the lack of object-oriented prior knowledge. The object appearance information is only available on-the-fly after the tracking process has already begun. To resolve this limited availability of target-oriented information, both generic appearance and motion features need to be integrated anonymously in the tracker, meanwhile the tracker's efficiency, adaptiveness, and generalization capability should be maintained.
\begin{figure}
\centering
\includegraphics[width=0.38\textwidth]{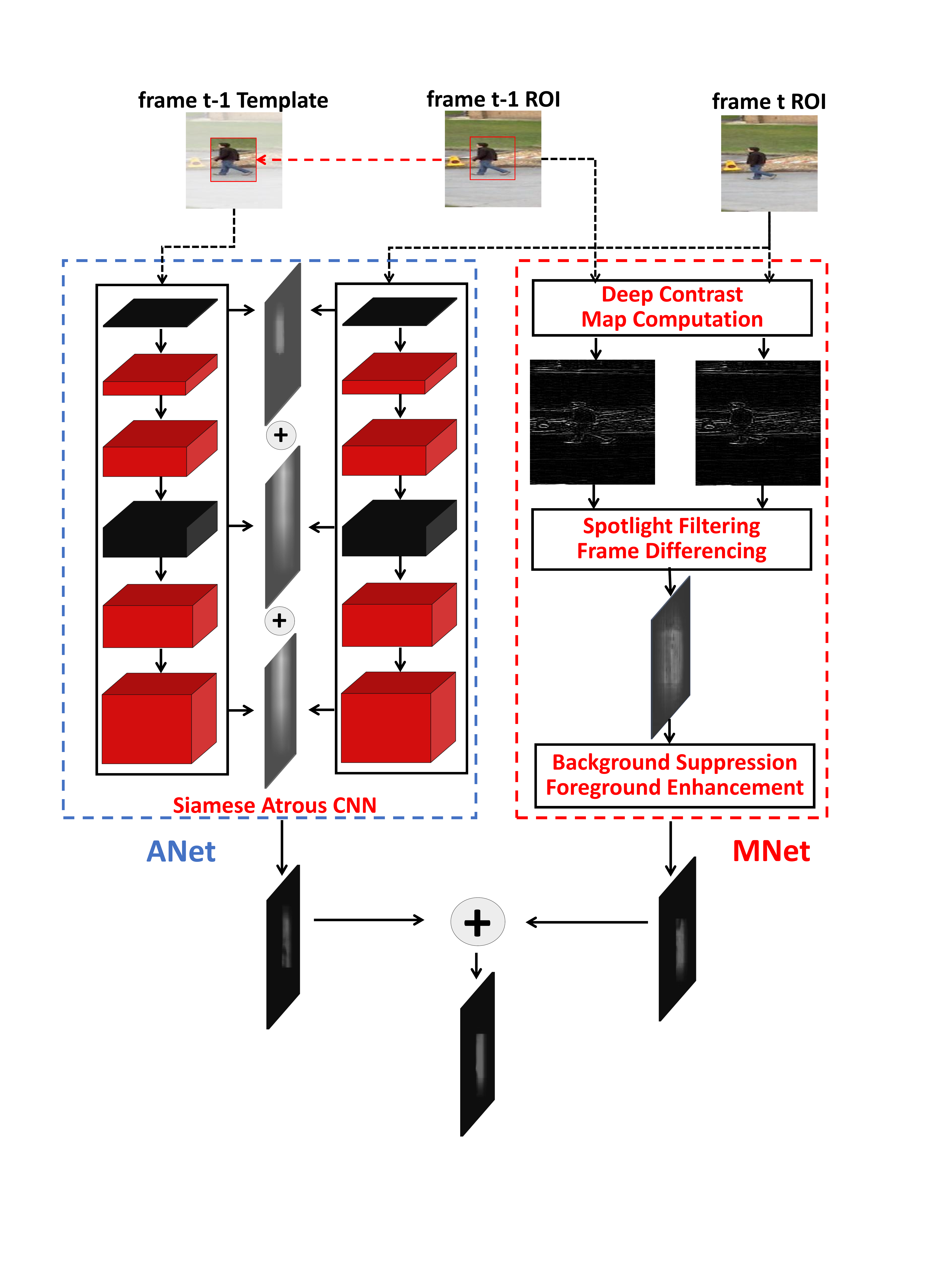}
\caption{Illustration of the proposed AMNet.}
\label{fig:WholeNet}
\end{figure}

Appearance features are deterministic image evidences indispensable to recognize and localize objects. Traditionally, object appearance is modelled with hand-crafted feature representation. CNNs have been developed and proven to be a more general and robust feature representation in various domains ~\cite{7,8}. But in the context of real-time tracking, establishing an online trained CNN-based appearance model is not feasible, considering the problem of training sample scarcity and prohibitive computation overhead. As a workaround, several trackers have been designed adopting the transfer learning strategy, i.e., extensively pre-trains CNN feature extractor on large offline datasets, then online fine-tunes the model to gain video-specific knowledge~\cite{FCNT, MDNet}. The popular Correlation Filter (CF) tracking paradigm also strives to learn an appearance model online~\cite{KCF, DSST, CCOT, ECO}. Due to the correlation-based one-stage localization and fast fourier domain computation, most CF trackers enjoy favorable tracking speed. To further enhance the real-time tracking performance, the offline trained Siamese trackers has raised increasing attentions~\cite{SiameseFC, CFNet, GOTURN, RASNet}. Instead of directly learning the generative appearance model of the object, Siamese trackers manage to learn an anonymous similarity embedding to achieve the tracking-by-matching strategy via comparing target candidates to the target template. Such a strategy is well-suited for model-free tracking without sacrificing the real-time performance.
{\setlength\abovedisplayskip{1pt}
\setlength\belowdisplayskip{1pt}
\begin{figure}
\centering
\includegraphics[width=3in]{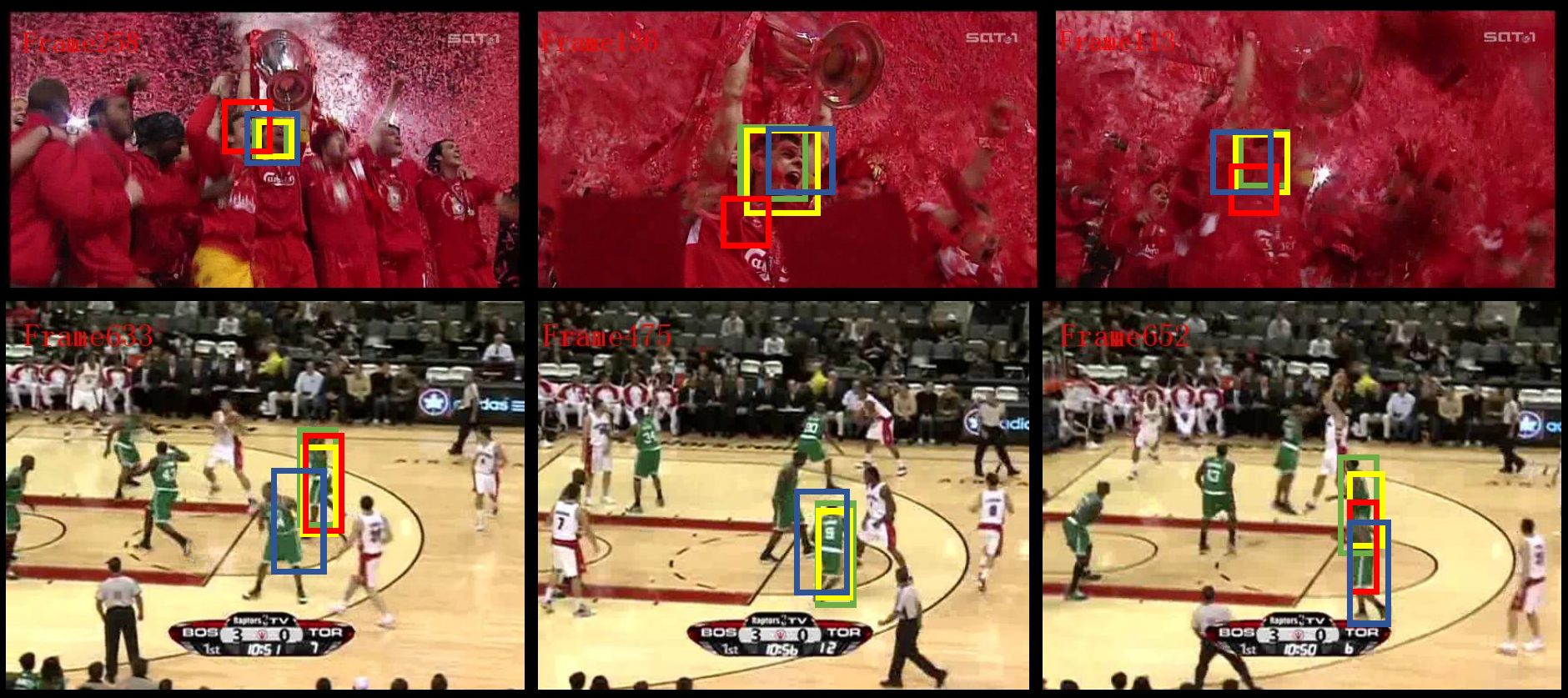}
\caption{Tracking results visualization of the proposed method comparing with states-of-the-art in appearance altering scenarios such as occlusion and deformation. AMNet, SiameFC, TCNN, and CCOT is marked with green, red, blue, and yellow bounding box respectively.}
\label{fig:bboxCompare}
\end{figure}

Motion features are helpful to facilitate model-free tracking, considering the fact that the motion of objects is pervasive and generic image evidence unrelated to specific object types. One approach to utilize motion information is by conducting motion predictions to generate Region-of-Interest (ROI). The other approach is to localize object with optical flow based methods~\cite{FlowNet, CVPR18E2EFlowCorrelation} or frame differencing based motion detections~\cite{FrameDIffShah, ShahECCV}. Between the two strategies, the frame differencing approach is more light-weight to compute, therefore it offers an efficient solution to integrate motion features. However, this approach more prone to be defected by motion noises so that it is not robust enough to perform tracking alone. Nonetheless, it can still offer localization information in form of a spatial attention mechanism. Although few attempts have been made, yet we believe it is contributive to introduce the feature encoding capability of general CNN in computing frame differencing motion detection. MNet is therefore designed as an end-to-end trainable CNN structure, which is able to generate generic motion detection in face of motion noises.

It is worth noticing that motion features have been under-emphasized in the community of object tracking. Most trackers focus on investigating the appearance features while keeping less attention regarding to the motion features. Nonetheless, we observe that appearance and motion are complementary and collaborative features in solving object tracking. Motion features are robust against appearance variations while the appearance features provide hard image evidences to correct faulty motion detections (as demonstrated in Figure \ref{fig:bboxCompare}, the proposed method shows better results in scenes with appearance variations). Consequently, both appearance and motion features should be jointly explored in model-free tracking methods.

Following the aforementioned motivations, in this work we design the AMNet to collaboratively integrate appearance and motion features to facilitate model-free tracking in an end-to-end trained network. As shown in Figure \ref{fig:WholeNet}, AMNet is composed with both ANet and MNet as two streams, where appearance and motion features are processed in parallel and each results in a localization response map. Specifically, ANet establishes a multi-scale filtering-based tracking-by-matching framework within a Siamese atrous CNN. Being essentially a similarity embedding network, the ANet is well-capable to achieve model-free tracking with great generalization capability. By progressively updating the object template via substituting the template using the newly tracked target state, the adaptiveness of ANet is also well-founded. MNet takes in two ROI patches of consecutive frames and outputs the response map by performing deep frame differencing motion detection. MNet contains the deep contrast map computation, Spotlight Filtering frame differencing, as well as the background suppression and foreground enhancement modules. All modules in MNet are designed to be lightweight, managing to integrate motion features as a spatial attention mechanism with acceptable computation overhead. The final tracking result is generated by concatenating the two response maps from each sub-network in depth then applying a 1 $\times$ 1 convolution layer to output a depth 1 map. The offline trained AMNet only takes one forward pass at each frame in test time, thus ensuring favorable efficiency performance.

The remainder of the paper is organized as follows: Section~\ref{sec:RW} provides a review of methods related to our work, Section~\ref{sec:AMNet} describes the proposed AMNet in-depth, experimental results and evaluation are demonstrated in Section~\ref{sec:Experiments}. We conclude the paper in Section~\ref{sec:Conclusions}.
\vspace{-1mm}
\section{Related Work}
\label{sec:RW}
Comprehensive and general surveys about object tracking please see ~\cite{SurveyTwo}. In this section we only focus on reviewing the methods that are closely related to the proposed approach.

\subsection{Appearance features in Correlation Filter based trackers}
The major concerns of applying online maintained CNN trackers are the sample scarcity and heavy computation overhead. One solution to lift these limits is the Correlation Filter based tracker. Essentially, CF localizes the target via exhaustively filtering through a region by computing cross correlations. Thanks to the efficient operations in the Fourier domain and the circulant dense sampling ability, CF is well qualified to handle the sample scarcity and efficiency issues. CF is also practicable to be extended to online CNN tracking, integrating the inherent CNN feature hierarchy to enable tracking on multiple scales ~\cite{HCF,Hedged}. This deep feature hierarchy is also integrated in continuous spatial domain via implicit interpolation model in C-COT~\cite{CCOT} and ECO~\cite{ECO}, resulting in continuous convolution filters. In the CREST tracker~\cite{CREST}, CF is reformulated as a convolution layer and embedded into an end-to-end residual learning framework. The pyramidal deep feature hierarchy of CNN together with the filtering style correlation computation is a valuable structure for tracking object with scale-awareness~\cite{Dsiam}.

\subsection{Appearance features in Siamese-based trackers}
Siamese-based trackers strive to learn an embedding, not just a feature representation, to calculate the appearance similarity between the template and a target candidate. In this way, the appearance extractor and discriminator are integrated and trained compactly end-to-end, so that they can co-adapt and cooperate with each other. Some representative methods include the GOTURN~\cite{GOTURN}, SINT~\cite{SINT}, SiameseFC~\cite{SiameseFC}, CFNet~\cite{CFNet}, and DCFNet~\cite{DCFNet} trackers. The GOTURN tracker fuses the two streams with fully connected layers and regresses to the tracked bounding box; Differently, the SiameseFC tracker deploys a fully convolution structure, where the tracking result is provided as a response map by filtering the feature map of the object template patch through the ROI; CFNet is founded on the SiameseFC tracker, but is improved to formulate the plain cross-correlation similarity computation into a trainable CF layer. Aside from these methods, Siamese CNN structure is also readily to be extended into a pyramidal form, achieving the ability to conduct multi-scale feature extracting and matching. The SA-Siam tracker ~\cite{SASiam} deliberatively trains S-net and A-net in parallel to process appearance features from higher semantic and lower pixel levels. Moreover, in DSiam tracker~\cite{Dsiam} and also in the proposed method, this idea is implemented and resulting in favorable outcomes.

\subsection{Motion features in motion detection trackers}
Motion detection is a generic way to localize anonymous moving objects. Frame differencing~\cite{FrameDIffShah} and background subtraction~\cite{BGSub} are two of the more primitive strategies in this domain. A few primitive attempts have been made to solve motion detection with simple CNNs~\cite{DeepBG,DeepBGtwo}. Aside from these motion detection methods, optical flow has long been recognized as a helpful feature to extract motion cues~\cite{learningByTracking, CVPR18E2EFlowCorrelation, SINT}. However, the time expense of computing optical flow features is prohibitive comparing to the simple motion detection methods. Nevertheless, the above mentioned motion detection approaches seldom take the factor of camera motions into account, which are a main source of motion noises. In the proposed MNet, we establish an end-to-end trained network that is able to perform reliable motion detection in the presence of such motion noises. By its design, MNet can be recognized as a combination of frame differencing and noise suppression modules.

\section{Model-free Object Tracking with the AMNet}
\label{sec:AMNet}
As shown in Figure \ref{fig:WholeNet}, the AMNet takes in two same size ROI patches ${z_{t}}$, ${z_{t-1}}$ on consecutive frames and one object template patch ${x_{t-1}}$ as input. All patches are cropped and resized on-the-fly in test time. The template patch is updated at each frame by replacing with the newly tracked target state in the previous frame. The ROIs are cropped from the same bounding box coordinates, centering at the object tracked location in the previous frame, with the size to be three times as that of the template. Based on the motion smoothness assumption, the object in the current frame should be enclosed in the ROI. Therefore, the tracking problem is reformulated to as fine-localizing the object inside the ROI. Such localization relies on the appearance information contained in the template patch, as well as the motion information resides in both ROI patches. Taking in the patches, ANet and MNet are two sub-networks processing appearance and motion features in parallel to result in response maps with the same size as the ROI. The final tracking result is a fusion of these two response maps by first concatenating them in depth, then applying a 1 $\times$ 1 convolution layer to generate an output map with depth of 1. Noteworthily, a similar dual stream deep convolutional network structures has been proposed to integrate appearance and optical flow motion features to solve tracking in~\cite{AddOn}, which can be recognized as the first attempt in this line of research. However, this method only offers a rather preliminary solution, where both appearance and motion features are processed with pre-trained simple CNN structures, and the tracking is resolved as a two-stage foreground-background classification via a SVM classifier.

\subsection{ANet with Appearance Features}
Taking in the object template and the ROI, ANet is designed to localize the object within the ROI using a tracking-by-matching strategy, i.e., searching through the ROI to highlight the region which is most similar to the object template. Two different approaches have been practiced to achieve tracking-by-matching. The discrete proposal-based methods sparsely sample candidates from the ROI, and then apply a trained classifier to compute a similarity score for each proposal~\cite{MDNet, TSN}. The efficiency of these methods is inferior as the repetitive operations on each proposal can be time-consuming. The other approach is to perform a continuous filtering search densely across the ROI in a sliding-window way. A similarity score is computed at each filtering location via calculating the cross-correlation between the template and the sampled patch. By reformulating the cross-correlation computation into a convolution and using the template as filter, this approach is well-suited to be integrated into a deep CNN framework to realize end-to-end trainable tracking-by-matching. Siamese CNN is therefore deployed to simultaneously pre-process both the template and the ROI patches into feature maps before matching. According to the formulation of the template, methods can be further categorized into the Correlation Filter based ~\cite{HCF, CCOT, CFNet, DCFNet, Dsiam} and the plain feature map filter based ~\cite{SiameseFC} categories. The Siamese structure obtained in these methods is essentially a model-free similarity embedding which is trained to establish matching between two image instances.
{\setlength\abovedisplayskip{1pt}
\setlength\belowdisplayskip{1pt}
\begin{figure}
\centering
\includegraphics[width=0.45\textwidth]{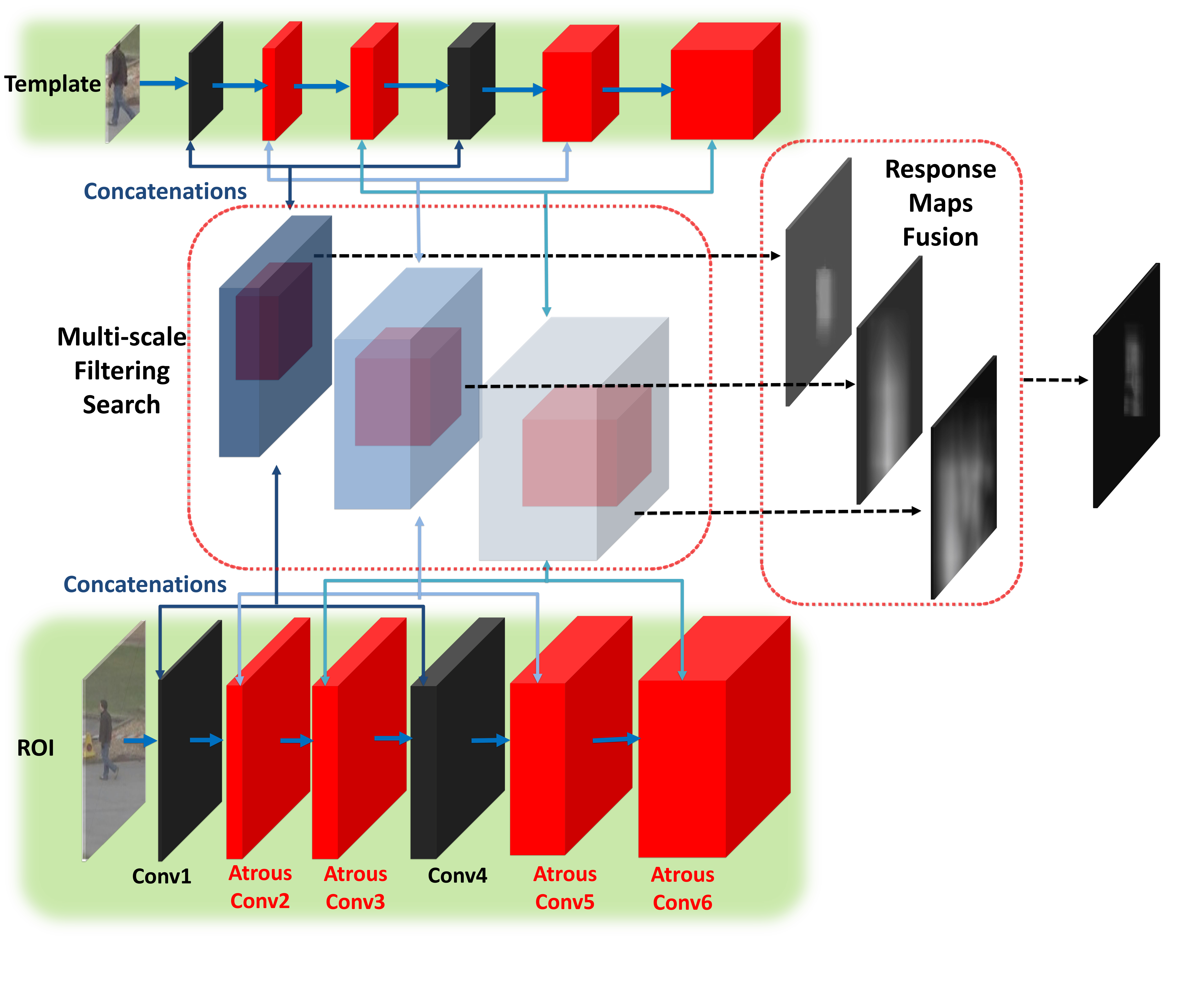}
\caption{Illustration of the ANet. The Siamese atrous CNN networks are marked in dark background, where atrous convolution layers feature maps are in red color.}
\label{fig:ANet}
\end{figure}

The proposed ANet follows the strategy of plain feature map filtering. Being entirely trained offline, the Siamese network is very efficient to test by simply performing a single forward pass at each frame. Moreover, because the plain CNN feature maps are used as filters, ANet further saves the online computation time which is otherwise consumed in maintaining the Correlation Filter updates. Besides the favorable efficiency, ANet also enables multi-scale tracking with the inherent CNN feature hierarchy. Different from the pyramidal feature hierarchy utilized in ~\cite{HCF, CCOT, Dsiam}, where pooling layers are interleaved to shrink the resolution so to enlarge the receptive field, in ANet we apply atrous convolutions with different atrous rates to establish the hierarchy. The advantage of adopting atrous convolutions in ANet is two-fold: firstly, the spatial resolutions of the feature maps are kept constant by removing the down-sampling as well as up-sampling layers, therefore contributing to finer pixel-wise localization; Secondly, atrous convolutions can maintain the feature spatial structure while removing feature redundancy. Furthermore, in order to avoid the gridding effect, in ANet we instantiate the atrous convolutions in a Hybrid Dilated Convolution (HDC) structure with convolution layers of various atrous rate.
{\setlength\abovedisplayskip{1.0pt}
\setlength\belowdisplayskip{1.0pt}
\begin{table*}[htb]
\renewcommand{\arraystretch}{1.0}
\caption{Architecture of the ANet}
\label{table1}
\centering
\footnotesize
\begin{tabular}{c c c c c c|}
\hline
\textbf{Layer} &\textbf{Filter Size} &\textbf{Output Channels} &\textbf{atrous Rate} &\textbf{Lateral Connection}\\
\hline
\textbf{Conv1} &5*5 &6 &1 &Conv4 \\
\textbf{Atrous Conv2} &3*3 &12 &2 &Atrous Conv5 \\
\textbf{Atrous Conv3} &3*3 &24 &3 &Atrous Conv6 \\
\textbf{Conv4} &5*5 &36 &1 &Conv1 \\
\textbf{Atrous Conv5} &3*3 &48 &2 &Atrous Conv2 \\
\textbf{Atrous Conv6} &3*3 &64 &3 &Atrous Conv3 \\
\hline
\end{tabular}
\end{table*}

Viewing horizontally in Figure \ref{fig:ANet}, the template ${x_{t-1}}$ and ROI ${z_{t}}$ patches are parallelly processed by the parameter-sharing branches of the Siamese atrous CNN, generating feature maps ${X}$ and ${Z}$ (the subscript is eliminated for clarity) with increasing depth containing more semantic information:
{\setlength\abovedisplayskip{1.5pt}
\setlength\belowdisplayskip{1.5pt}
\begin{equation}
\footnotesize
{Z ^l} = {\phi ^l}({z_{t}}), {X ^l} = {\phi ^l}({x_{t-1}})
\end{equation}
where ${\phi(*)}$ indicates the convolutional embedding function, and upper script $l$ denotes the feature level in the hierarchy. Features at the same level of the two branches in the Siamese networks are equally embedded, i.e., represented under the same level of abstraction and transformation, and therefore the multi-scale filtering established between the two branches is reasonable. Before performing the filtering search, lateral connections are established by concatenating feature maps from different convolution layers. The concatenated ${Z ^L}$ and ${X ^L}$ contains richer cross-resolution features and are applied in the following filtering search:
{\setlength\abovedisplayskip{1.5pt}
\setlength\belowdisplayskip{1.5pt}
\begin{equation}
\footnotesize
{O^L} = f({Z^L},{X^L}) = {Z^L} \otimes {X^L}
\end{equation}
${f(*)}$ is the embedded cross-correlation layer generating scalar value similarity score map ${O^L}$ (no padding added to avoid marginal localization ambiguity). This layer is non-parameterized and back propagate friendly, which can be easily integrated into a CNN structure. The final output response map ${O_A}$ is computed by first stacking all ${O^L}$ in depth then apply a $1 \times 1$ convolution to output a depth 1 feature map. The detailed architecture of ANet is shown in Table \ref{table1}.

\subsection{MNet with Motion Features}
\label{sec:MNet}
As shown in Figure \ref{fig:MNet}, the fully convolutional MNet consists of three components. Taking in two overlapping ROIs cropped from consecutive frames, MNet is designed to output a same size motion detection response map depicting the moving regions. By highlighting the moving foreground object, this response of MNet provides generic spatial attention contributing to the localization of object without introducing extensive computation overhead. To our knowledge, we are among the first to attempt solving frame differencing motion detection within an end-to-end trainable CNN framework. It is non-trivial to achieve such a solution when camera motion is present, which extensively contaminate the foreground motion response with noises. Traditionally, image registration is required as a pre-processing step to counteract the moving camera. While in the context of MNet, all modules are deliberatively designed to suppress the noise while enhancing the foreground response to cancel out the camera motion. Noteworthily, the frame differencing motion detection instantiated in MNet is fully generic, well-capable to facilitate model-free tracking of any moving targets.
\vspace{-3mm}
\paragraph{Deep Contrast Computation Module.}
Previous work has shown that comparing to directly subtracting a pair of original images, computing frame differencing with contrast maps are more robust to motion noises \cite{DetectionContrast, FrameDIFFnew, FrameDIffShah}, as the pervasive background motion can be evened out by the gradient-like computation considerably. Essentially, the computation of contrast maps is carried out by filtering the image with a pre-defined filter (e.g. Laplacian or Sobel kernels). In this proposed module, we strive to learn such filters adaptively. In particular, filters of different sizes are cascaded to gain sensitivity to contrast regions with different sizes. Illustrated in Figure \ref{fig:MNet}, for implementation we use three convolution layers with Relu activations but no interleaved pooling layers to form the contrast computation module. All convolution filters are with stride 1 and of the same input-output channels, resulting in contrast maps ${Z^{C}_{t}}$, ${Z^{C}_{t-1}}$ who have the same size and dimension as the input ROIs (${z_{t}}$, ${z_{t-1}}$). The computation of this module can be formulated as:
{\setlength\abovedisplayskip{1.5pt}
\setlength\belowdisplayskip{1.5pt}
\begin{equation}
\begin{aligned}
\footnotesize
{Z^{C}} = {\psi}({z})
\end{aligned}
\end{equation}
where ${\psi(*)}$ is the joint deep embedding of all convolution layers. For one layer (take conv3 for instance), the computation at the $j$th filtering location can be specified as:
{\setlength\abovedisplayskip{1.5pt}
\setlength\belowdisplayskip{1.5pt}
\begin{equation}
\begin{aligned}
\footnotesize
{Z_3}^{C}_{,j} = {F_3}(z) = \sum\limits_{i = 1}^{3 \times 3} {{f_j^{i}}*{z_j^{i}}}
\end{aligned}
\end{equation}
${F_3(*)}$ denotes the filtering operation of conv3, where ${f_i}$ is the $i$th element in the filter.

{\setlength\abovedisplayskip{1pt}
\setlength\belowdisplayskip{1pt}
\begin{figure*}
\centering
\includegraphics[width=0.78\textwidth]{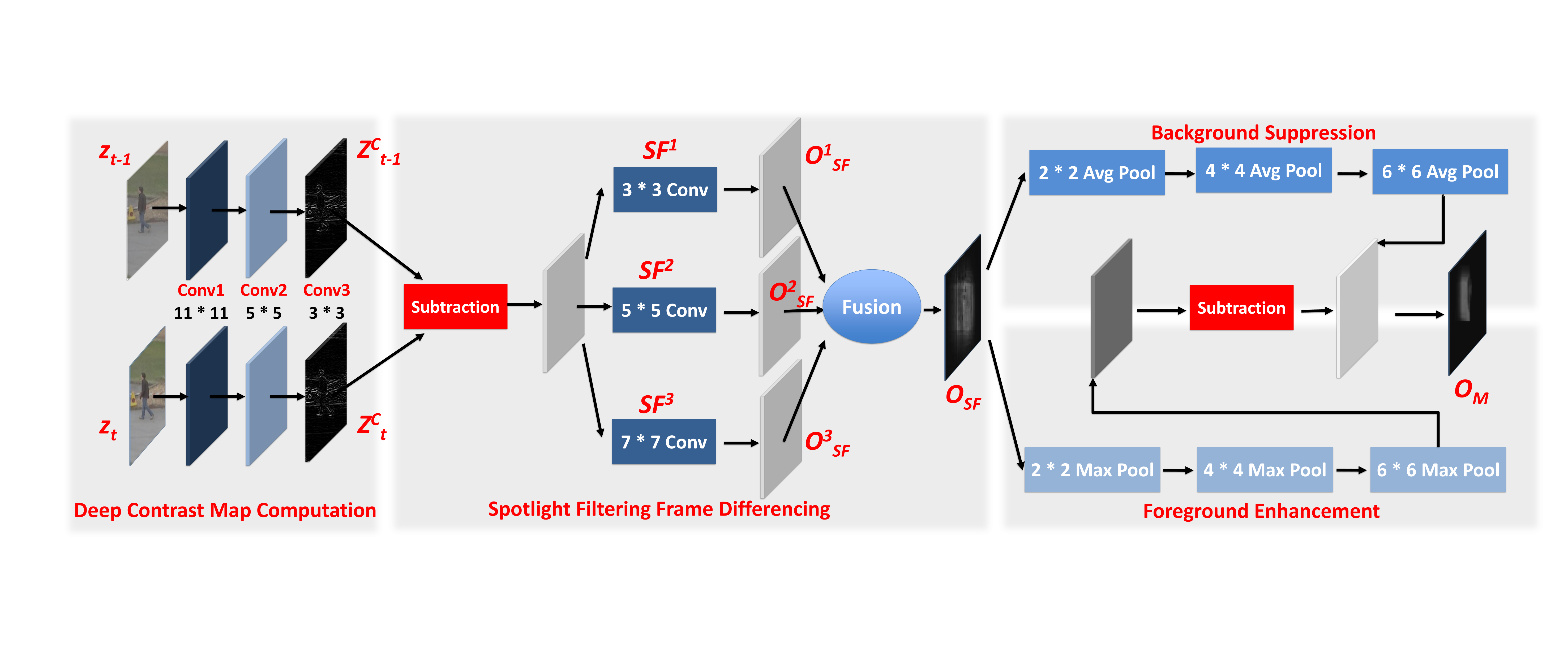}
\caption{Illustration of the MNet.}
\label{fig:MNet}
\end{figure*}
\vspace{-3mm}
\paragraph{Spotlight Filtering Frame Differencing Module.}
The concept of Spotlight Filtering is implemented to generate frame differencing response with a pair of contrast maps. Instead of element-wise subtractions, the Spotlight Filtering module reformats the computation to patch-wise subtraction. The module starts with aligning two contrast maps ${Z^{C}_{t}}$ and ${Z^{C}_{t-1}}$ spatially, and then conducting element-wise subtraction within a pair of aligned patches with the same size as the Spotlight Filter $SF \in {R^{k \times k}}$. The computed response at a given filtering region is the summation of all the element-wise subtractions within the region. We formally describe the computation at the $j$th filtering location between two contrast maps $A$ and $B$ as:

{\setlength\abovedisplayskip{1pt}
\setlength\belowdisplayskip{1pt}
\begin{equation}
\begin{aligned}
\footnotesize
S{F_j}(A,B) = \sum\limits_{i = 1}^{k \times k} {(A_j^i - B_j^i)}
\end{aligned}
\end{equation}
Accordingly, the Spotlight Filtering response map $O_{SF}^l$ between ${Z^{C}_{t}}$ and ${Z^{C}_{t-1}}$ is computed as:
{\setlength\abovedisplayskip{1pt}
\setlength\belowdisplayskip{1pt}
\begin{equation}
\begin{aligned}
\footnotesize
O_{SF}^l({Z^{C}_{t}},{Z^{C}_{t-1}}) = S{F^l}({Z^{C}_{t}},{Z^{C}_{t-1}})
\end{aligned}
\end{equation}
where $l$ indicates Spotlight filters with different sizes. This multi-scale structure is applied to provide different receptive fields to adapt to motion with varied magnitudes. Response maps from different scales are fused depth-wise with a 1 $\times$ 1 convolution layer, resulting in the Spotlight Filtering response map $O_{SF}$. The design of Spotlight Filtering is resilient to camera motions, as the level of spatial abstraction introduced by the filters' receptive fields dilutes the image-level motion noise. To instantiate the Spotlight Filtering module with simple fully-convolution networks, as demonstrated in Figure \ref{fig:Spotlight}, it first computes the element-wise subtraction between two inputs, and then applies different size convolution filters parallelly on the output subtracted feature maps. In this way, the $SF(*)$ filtering is further extended into a weighted version, where $W \in {R^{k \times k}}$ is the applied convolution filter:
{\setlength\abovedisplayskip{1pt}
\setlength\belowdisplayskip{1pt}
\begin{equation}
\begin{aligned}
\footnotesize
S{F_j}(A,B) = \sum\limits_{i = 1}^{k \times k} {{W^i}*(A_j^i - B_j^i)}
\end{aligned}
\end{equation}

\paragraph{Background Suppression and Foreground Enhancement Module.} The response map generated from the Spotlight Filtering module may still be contaminated by noises. This suppression and enhancement module is a hierarchical pooling structure to finalize the response map $O_M$. In particular, the background suppression module consists of three cascade average pooling layers. The pooling layers are all with stride 2, but with different kernel sizes to be selectively responsive to motions with different magnitudes. The foreground enhancement module is similarly configured, but with max pooling layers. Each kernel offers a level of abstraction, while the enhancement module will output a response map highlighting the dominating foreground motion, by subtracting it with the suppression module response map, this module works similar to a median image background subtraction operation \cite{ShahECCV}, which can further counteract with the camera motions. At the meantime, we observe this module is also morphologically effective to clean up the response map. The final response map of AMNet $O_{AM}$ is the deep fusion of $O_A$ and $O_M$ generated from two sub-networks.

{\setlength\abovedisplayskip{1pt}
\setlength\belowdisplayskip{1pt}
\begin{figure}
\centering
\includegraphics[width=0.16\textwidth]{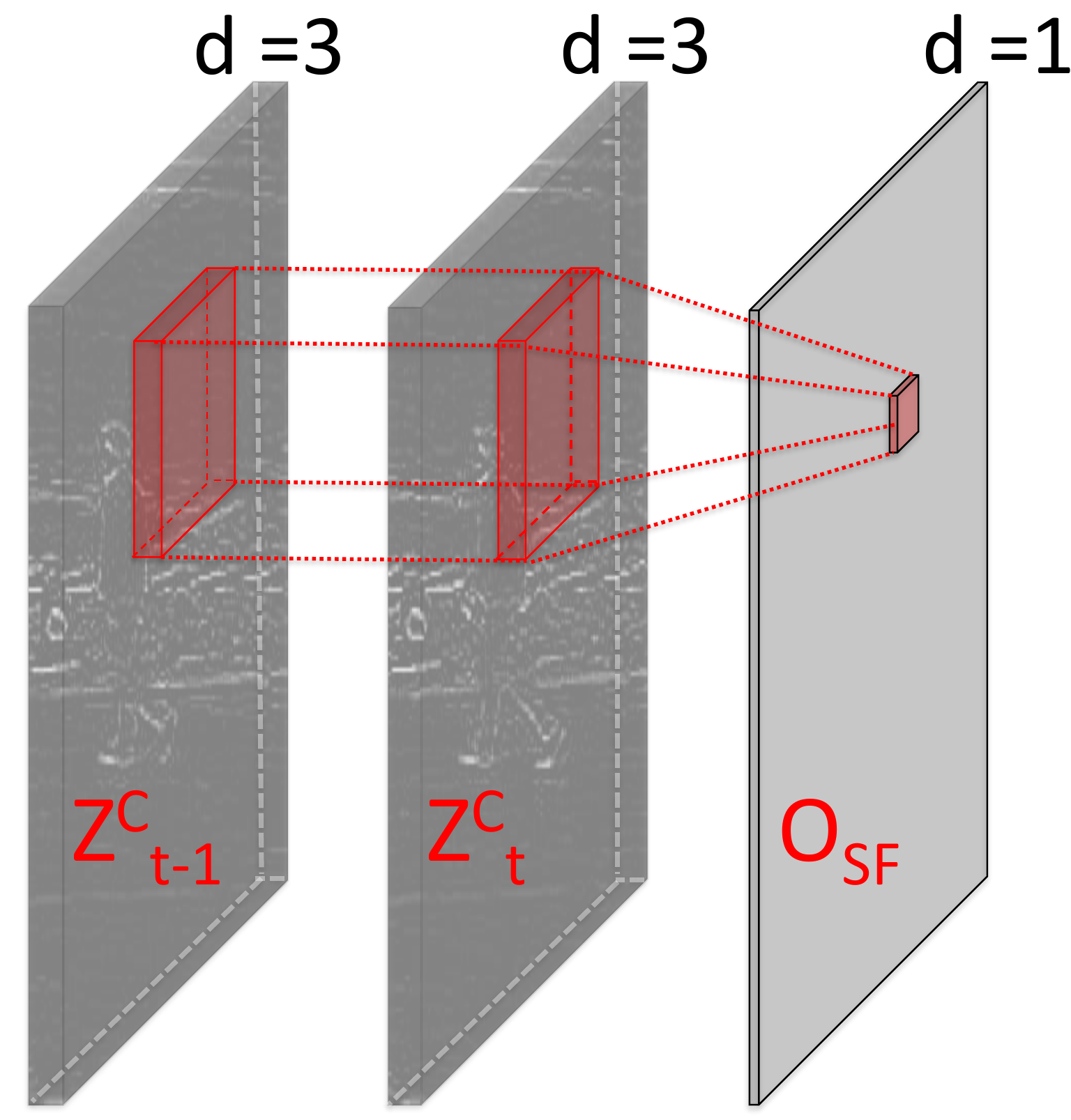}
\caption{Illustration of the Spotlight Filtering module.}
\label{fig:Spotlight}
\end{figure}
\vspace{-0.5mm}
\section{Experiments}
\label{sec:Experiments}
\vspace{-0.5mm}
The proposed method is implemented in Python using the Tensorflow deep learning platform. Experiments are conducted on the OTB2013~\cite{OTB2013}, TB50~\cite{OTB}, and VOT2017 datasets. 11 states-of-the-art (ECO\cite{ECO}, CCOT \cite{CCOT}, SINT \cite{SINT}, CREST \cite{CREST}, SimaeseFC \cite{SiameseFC}, CFNet\cite{CFNet}, Struck \cite{Struck}, HCF \cite{HCF}, TLD \cite{TLD}, IVT\cite{IVT}, CSK\cite{KCF}) participate in the comparison on OTB2013 and TB50 benchmarks, where the performances are measured with the one-pass evaluation using precision and success plot. On VOT2017, we compare with 14 published trackers (CCOT \cite{CCOT}, Staple \cite{STAPLE}, GMDNetN \cite{MDNet}, SiamFC \cite{SiameseFC}, ASMS \cite{ASMS}, DSST \cite{DSST}, UCT, SiamDCF, CMT \cite{CMT}, ECO, CFCF \cite{CFCF}, LSART \cite{LSART}, GMD, ANT \cite{ANT}), where the expected average overlap (EAO), accuracy, and robustness metrics are applied as the measurements.

\subsection{Tracking Methodology}
Tracking with AMNet is performed iteratively frame-by-frame. Based on the tracked object state at frame t-1, moving onto frame t three patches ${z_{t}}$, ${z_{t-1}}$ and ${x_{t-1}}$ are cropped and resized. In the experiments templates are resized to 64 ${\times}$ 64, and ROIs are resized to 192 ${\times}$ 192. Particularly, the target template is updated progressively via replacing with the newly tracked state in the previous frame. Based on the observation that appearance variations of targets usually progress gradually along with the elapsing frames, so that the inter-frame changes of targets are not radical, therefore such a progressive update scheme holds favorable balance between robustness and adaptiveness performance. By feeding all three patches into the trained AMNet, one forward pass is executed to generate the response map. The object state at frame t is estimated by searching the maximum value on the map into a bounding box annotation, upon which the tracking is extended onto frame t+1.

\subsection{AMNet Training}
Tracking by essence is a localization task where spatial resolution matters, and therefore AMNet maintains rather shallow configurations without down-sampling to maintain the spatial information. For this reason, AMNet is end-to-end trained from scratch but not fine-tuned on pre-trained CNN structures such as AlexNet or VGG. Training data are triplets of image patches, including two ROIs and one template. Patches are cropped and resized from a combination of NUS-PRO~\cite{NusPro}, TempleColor128~\cite{TempleColor}, and MOT2015 datasets~\cite{MOT}. Sequences overlapping with OTB and VOT are discarded.
{\setlength\abovedisplayskip{1.5pt}
\setlength\belowdisplayskip{1.5pt}
\begin{equation}
\footnotesize
L = {\sum\limits_j^{} {\left\| {O_{AM}^j - O_{gt}^j} \right\|} ^2} + {L_{reg}}
\end{equation}
The loss $L$ to be minimized in the training is an element-wise ridge loss computed between the predicted and ground truth response maps, and $j$ denotes all elements in the map. The regularization term ${L_{reg}}$ in the loss is achieved implicitly using the weight decay method. ${O_{AM}}$ values are squashed with a sigmoid activation. ${O_{gt}}$ is generated by placing a 2D Gaussian distribution peak at the ground truth bounding box location. During training we deployed Xavier initialization, Adam optimizer with weight decay of 0.005, learning rate starts at 1 * 10-3, and step-wisely drops to 1 * 10-5. Mini-batch size is set to 16.

\subsection{Results}
\paragraph{OTB2013 and TB50 Datasets.}
For comprehensiveness, comparison results are shown on both OTB2013 and TB50 benchmarks, where TB50 dataset is in general more challenging and representative than OTB2013. As shown in figure \ref{fig:OTB2013TB50}, the proposed method performs among the best in both metrics overall. In particular, examining closely w.r.t. other Siamese trackers, CFNet adds a correlation layer based on SiameseFC, but the performance gain is not significant. AMNet excels both of them by a considerable margin, thanks to the multi-scale atrous configuration as well as the integration of motion features. SINT tracker (which has only been evaluated on the OTB2013 dataset by the authors) shows competitive performance with AMNet considering the fact that it also integrates motion features via deploying optical flow. However, optical flow features are off-the-shelf so that SINT is not end-to-end trainable, and it runs significantly slower than AMNet. In comparisons with state-of-the-arts adopting multi-scale Correlation Filter with CNN feature hierarchy, HCF, CREST, CCOT and ECO exhibit similar or better accuracy performance, yet their efficiency performances are greatly limited by the time consuming online training and maintaining of CF, therefore all running at around 1 FPS. Although ECO speeds up CCOT to 8 FPS with the implementation of factorized convolution operators, yet they are all still much slower than AMNet.

{\setlength\abovedisplayskip{1.0pt}
\setlength\belowdisplayskip{1.0pt}
\begin{figure}
\centering
\subfigure[OTB2013]{
\label{fig:OTB2013TB50:a} 
\includegraphics[width=3in]{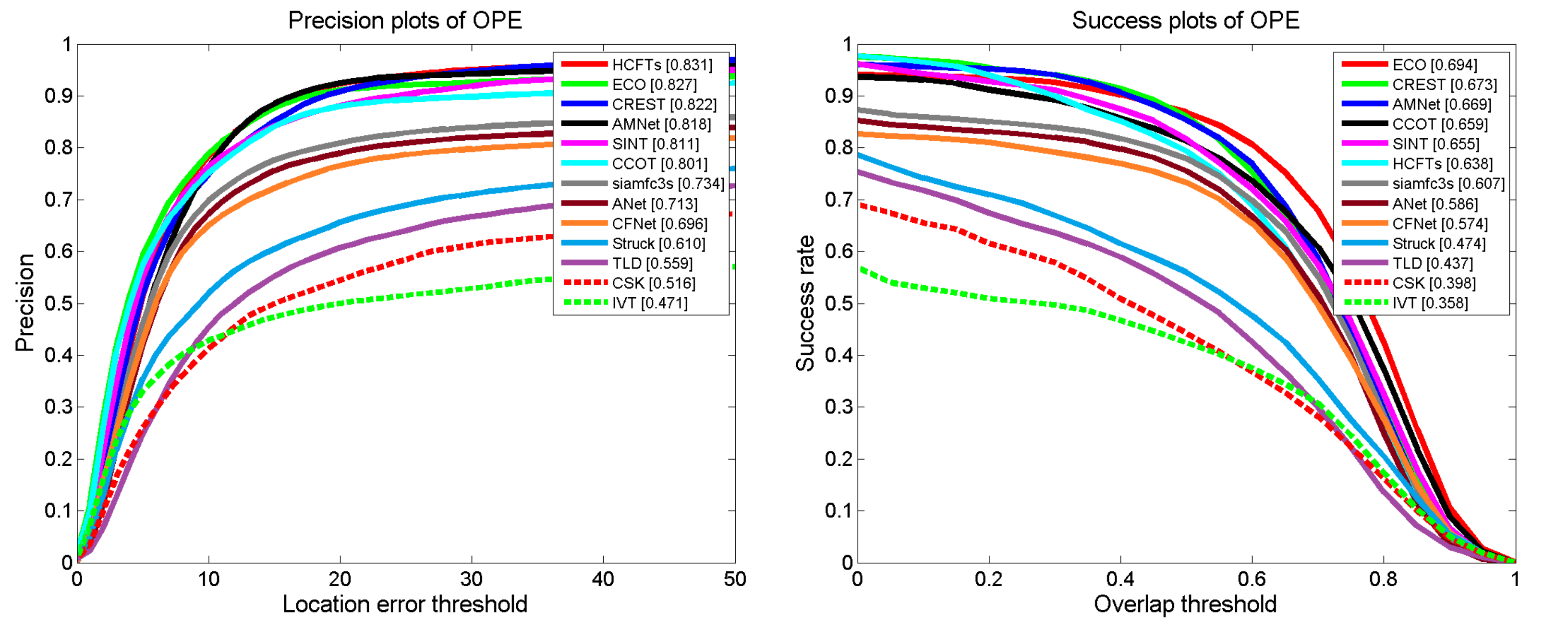}}
\hspace{0.1in}
\centering
\subfigure[TB50]{
\label{fig:OTB2013TB50:b} 
\includegraphics[width=3in]{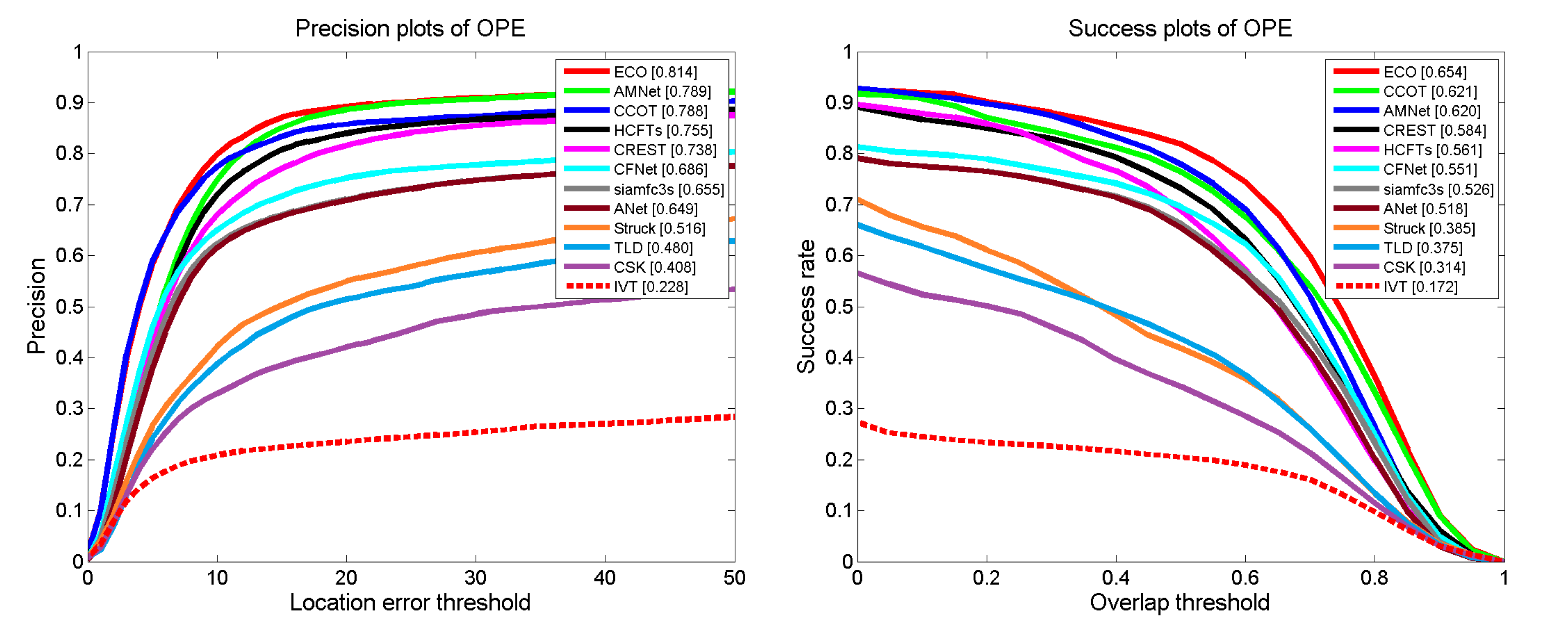}}
\hspace{0.1in}
\caption{Precision and success plots using OPE evaluation on OTB2013 and TB50 datasets.}
\label{fig:OTB2013TB50} 
\end{figure}

{\setlength\abovedisplayskip{1pt}
\setlength\belowdisplayskip{1pt}
\begin{figure}
\centering
\includegraphics[width=2.6in]{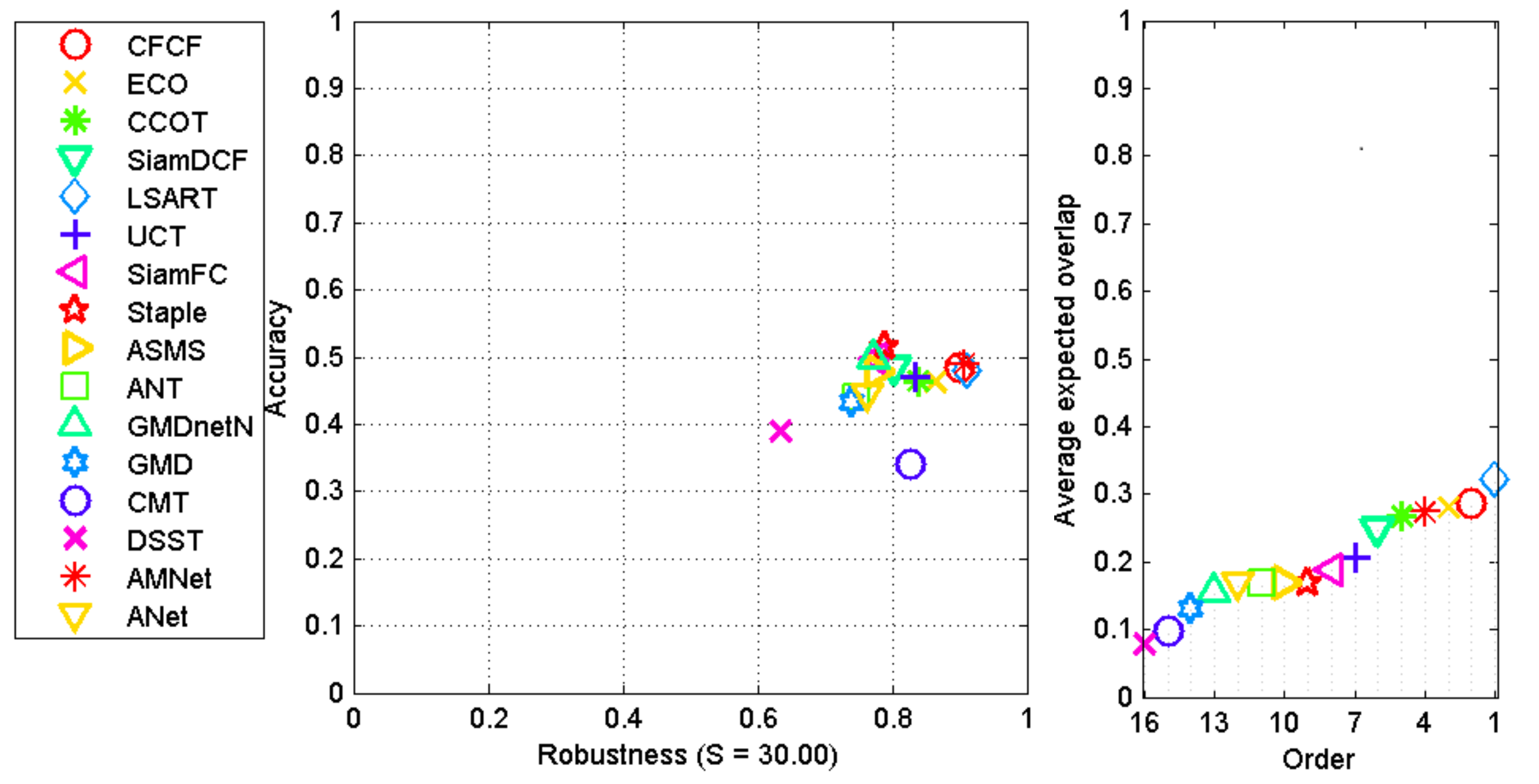}
\caption{EAO rankings and AR plot on VOT2017 dataset.}
\label{fig:VOT2017}
\vspace{-3mm}
\end{figure}

{\setlength\abovedisplayskip{1.0pt}
\setlength\belowdisplayskip{1.0pt}
\begin{figure*}
\centering
\subfigure[OTB2013]{
\label{fig:OTB2013TB50:a} 
\includegraphics[width=6.9in]{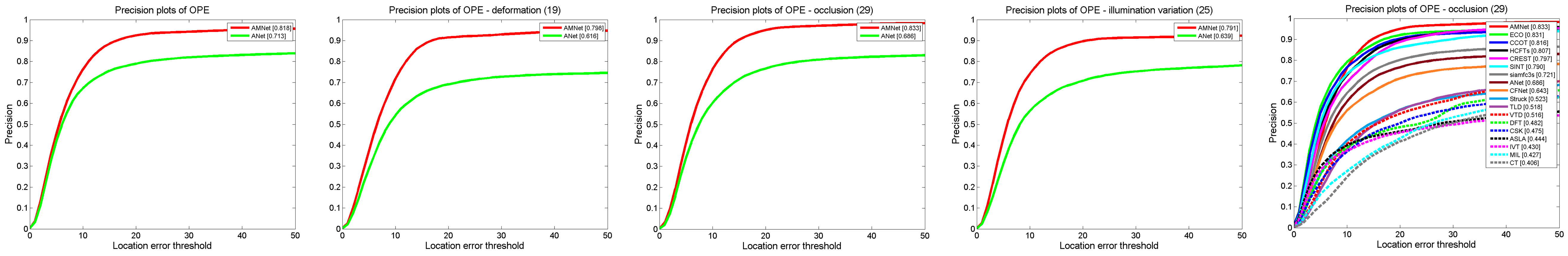}}
\hspace{0.1in}
\centering
\subfigure[VOT2017]{
\label{fig:OTB2013TB50:b} 
\includegraphics[width=4in]{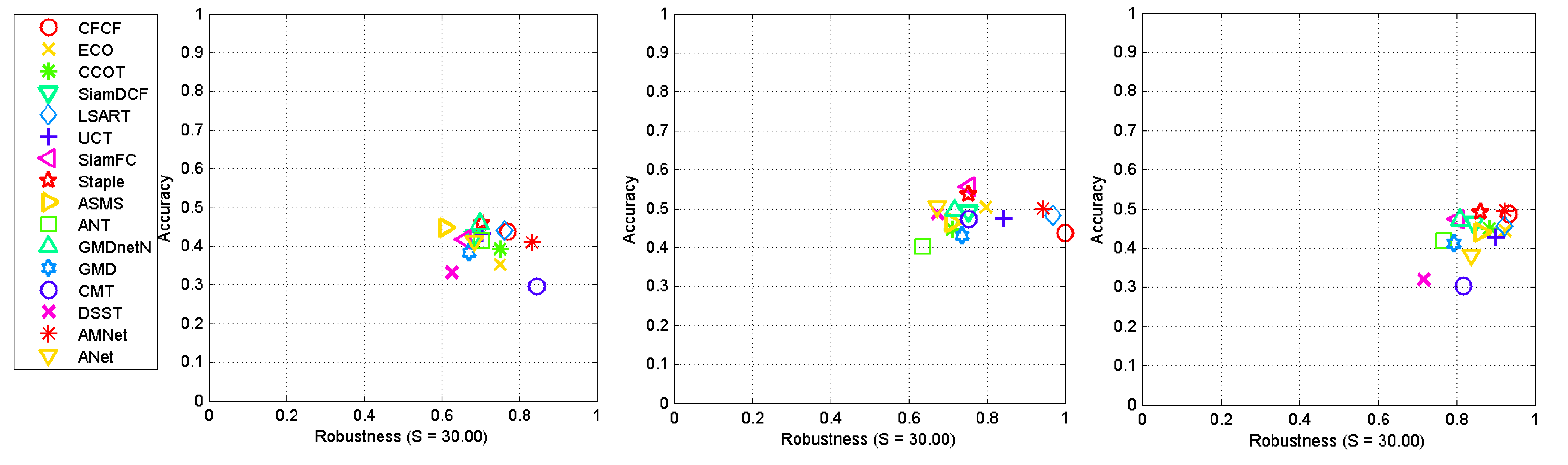}}
\hspace{0.1in}
\caption{Tracking performance comparisons in scenes with appearance variations on OTB2013 and VOT2017. In (b), sub-figures from left to right show the plots in sequences with occlusion, illumination, and size variations respectively.}
\label{fig:Ablation} 
\end{figure*}
\vspace{-4mm}
\paragraph{VOT2017 Dataset.}
As illustrated in Figure \ref{fig:VOT2017}, trackers are evaluated and compared on VOT2017 datasets with EAO ranking as well as Accuracy-Robustness (AR) plot. Reporting similar results as on the TB datasets, AMNet is also among the top-performers overall. Particularly, AMNet still outperforms SiamFC (i.e. the SiameseFC tracker) by a considerable margin. Besides, AMNet also performs better than SiamDCF, which is an improved version of SiameseFC by combining with the DCFNet tracker \cite{DCFNet}. GMD is a collaboration of GOTURN and MDNet trackers, realizing a timely feedback on model update for the underlying Siamese CNN tracker. Motion-wise, ANT and CMT apply optical flow to integrate motion features, AMNet surpasses both of them in tracking accuracy significantly, demonstrating the advantage of integrating appearance with motion features. For the state-of-the-art methods that out-rank AMNet in the EAO metric, LSART tracker learns a kernelized ridge regression together with spatially regularized convolutional neural networks to solve tracking, running at 1 FPS on a CPU. CFCF tracker, which is the winner on the general  VOT2017 challenge, integrates CNN features with the HOG and Colour Names (CN) features under a CF tracking framework, and can achieve leading tracking performance at around 2 FPS. Together with ECO, these three trackers beat us in EAO rankings, while the AMNet demonstrates higher tracking efficiency in comparison with all of them.

{\setlength\abovedisplayskip{0.9pt}
\setlength\belowdisplayskip{0.9pt}
\begin{table}[htb]
\renewcommand{\arraystretch}{1.1}
\caption{Performance of selected top-ranking trackers on OTB2013. First column lists the success plot AUC percentages. Statistics that outperform AMNet are marked in blue.}
\label{ExpDiscussion}
\centering
\scriptsize
\begin{tabular}{| c | c c |}
\hline
\textbf{Trackers} &\textbf{AUC} &\textbf{FPS}\\
\hline
\color{red}\textbf{AMNet} &\color{red}\textbf{0.669} &\color{red}\textbf{32}\\
\hline
\textbf{ECO} &\color{blue}0.694 &8\\
\textbf{SINT} &0.655 &4\\
\textbf{SiameseFC} &0.607 &\color{blue}86\\
\textbf{CFNet} &0.574 &\color{blue}75\\
\textbf{TCNN} &\color{blue}0.682 &2\\
\textbf{HCF} &0.638 &1\\
\textbf{CREST} &\color{blue}0.673 &1\\
\hline
\end{tabular}
\end{table}
\vspace{-2mm}
\paragraph{Ablation Study.}
Extended ablation studies are provided in this section to demonstrate the contribution of integrating motion features with appearance features. As mentioned in Section \ref{sec:Intro} and Section \ref{sec:MNet}, the MNet is proposed to robustly capture the frame differencing motion detection response, which is implemented to efficiently integrate motion features as a complimentary spatial attention mechanism to the appearance features. To elaborate on the contribution of integrating motion features, we hereby ablate the MNet and compare the tracking performance of solely running ANet with that of the complete AMNet. Being invariant to appearance variations, motion features are useful compliments in scenes with occlusion, deformation, scale and illumination variations, etc. This statement is exemplified in Figure \ref{fig:Ablation} (a). As shown on OTB2013 dataset, AMNet exceeds ANet in success rate AUC measure by 10.5\% overall, this improvement grows to 18.2\%, 14.7\%, and 15.2\% in scenarios of deformation, occlusion, and illumination variation, proving the fact that integrating motion features with appearance is beneficial for tracking, and this benefit magnifies in appearance-altering tracking scenes. Furthermore, the rightmost sub-figure illustrates that AMNet achieves the best success rate among all participants in scenes with occlusions, which further demonstrates the superiority of the deep integration of motion with appearance in AMNet. Above observations are also solidified on VOT2017 dataset as illustrated in Figure \ref{fig:Ablation} (b), where AMNet outperforms ANet significantly under challenging appearance-varying conditions.
\vspace{-2mm}
\paragraph{Discussion.} In the comparison, AMNet achieves the best performance jointly evaluating the tracking accuracy and efficiency. Listed in Table ~\ref{ExpDiscussion} are the top-performers in regard to tracking accuracy and speed, where AMNet, SiameseFC and CENet are the only three trackers who can reach real-time performance. AMNet leads the other two in AUC score by a large margin. Moreover, AMNet accuracy performance is on par with the best trackers overall.

This leading accuracy performance of AMNet can be attributed to the effective integration of appearance and motion features, resulting in more informative representation with robustness to appearance variations. The good efficiency of AMNet is realized by the offline training strategy, where the online computation expenses are relived at test time. What's more, all components in AMNet are deeply integrated and collaborated via end-to-end training, achieving an intact and cooperative solution. The model-free generalization capability of AMNet roots in the Siamese tracking-by-matching design of ANet, where scale-awareness is handled by the multi-scale pyramidal atrous CNN. Meanwhile, MNet by itself is also fully generic in processing motion detections. By progressively updating the object template with newly tracked target state, the adaptiveness of AMNet is improved. We hereby highlight that the major contribution in this work is the unified tracking framework that jointly explores both appearance and motion information.
\vspace{-1mm}
\section{Conclusions}
\label{sec:Conclusions}
AMNet is proposed in this work. By deeply integrating both appearance and motion features within an end-to-end trained framework, AMNet enables real-time model-free object tracking with favorable accuracy, efficiency, adaptiveness, and generalization performance. Specifically, we design ANet to realize filtering-based tracking-by-matching with a Multi-scale Siamese atrous CNN. In MNet, deep frame differencing motion detection with background suppression and foreground enhancement is achieved with robustness to camera motions. Extensive experiment results demonstrate the top overall performance of AMNet by jointly considering the tracking accuracy and speed.
\vspace{-1mm}
\section{Acknowledgement}
This paper was supported in part by the National Key Research and Development Program of China under Grant 2016YFB1200100,in part by the National Science Fund for Distinguished Young Scholars under Grant 61425014, in part by the National Natural Science of China under Grant 61871016.

{\small
\bibliographystyle{ieee}
\bibliography{Newegbib}
}

\end{document}